\documentclass{article} 
\usepackage[pdftex]{graphicx}

\usepackage{amsmath,amsfonts,bm}

\def\eqref#1{equation~\ref{#1}}

\def\1{\bm{1}}

\DeclareMathAlphabet{\mathsfit}{\encodingdefault}{\sfdefault}{m}{sl}
\SetMathAlphabet{\mathsfit}{bold}{\encodingdefault}{\sfdefault}{bx}{n}

\usepackage[a4paper, total={6.5in, 8in}]{geometry}
\usepackage{hyperref}
\usepackage{url}
\usepackage{array}

\title{Localization: A Missing Link in the Pipeline of Object Matching and Registration}

\author{Deepak Mishra$^1$, Rajeev Ranjan$^2$, Santanu Chaudhury$^1$, Mukul Sarkar$^1$\\ \& Arvinder Singh Soin$^3$ \\	
	$^1$Indian Institute of Technology Delhi, New Delhi, India\\
	$^2$Chennai Mathematical Institute, Chennai, India\\
	$^3$Medanta Hospital, Gurgaon, India\\
	\text{\{deemishra21@gmail.com\}}
}
\date{}

\begin{document}

\maketitle

\begin{abstract}

Image registration is a process of aligning two or more images of same objects using geometric transformation. Most of the existing approaches work on the assumption of location invariance. These approaches require object-centric images to perform matching. Further, in absence of intensity level symmetry between the corresponding points in two images, the learning based registration approaches rely on synthetic deformations, which often fail in real scenarios. To address these issues, a combination of convolutional neural
networks (CNNs) to perform the desired registration is developed in this work. The complete
objective is divided into three sub-objectives: object localization, segmentation and matching transformation. Object localization step establishes an initial correspondence between the images. A modified version of single shot multi-box detector is used for this purpose. The detected region is cropped to make the images object-centric. Subsequently, the objects are segmented and matched using a spatial transformer network
employing thin plate spline deformation. Initial experiments on MNIST and Caltech-101 datasets show that
the proposed model is able to produce accurate matching. Quantitative evaluation performed using dice coefficient (DC) and mean intersection over union (mIoU) show that proposed method results in the values of 79\% and 66\%, respectively for MNIST dataset and the values of 94\% and 90\%, respectively for Caltech-101 dataset.
The proposed framework is extended to the registration of CT and US
images, which is free from any data specific assumptions and has better
generalization capability as compared to the existing rule-based/classical
approaches.

\end{abstract}


\section{Introduction}

Object matching and image registration have gained considerable importance in recent years. It requires simultaneous processing of multiple images to establish pixel/voxel correspondences in presence of view point variations, modality change and variations in object characteristics. 
Image registration helps to process complementary information for applications like, image completion, scene recognition, 3D modeling, and navigation \cite{barnes2010generalized,nam2011automatic}. 

In general registration is formulated as a problem with a fixed and a moving image in which, latter is transformed to establish a correspondence between the two. Traditional approaches rely on local correspondences of pixel intensities and optimize a matching criteria \cite{ceritoglu2013computational,ashburner2007fast,revaud2016deepmatching}. The approaches are slow and memory intensive \cite{rohe2017svf}. Another common problem of the existing approaches is the assumption of the availability of object-centric images. Often comes the situation, where moving image contains the object of interest which needs to be located in fixed image before performing the matching operation. Moreover, the objects of interests are sometimes surrounded with similar objects. However, unlike the real-world scenario, the approaches mostly consider the images with focused single object. \cite{rocco2017convolutional}. 

In this work, we propose a novel method for object matching. It is a combination of convolutional neural networks (CNNs) which sequentially perform object localization, segmentation and matching. We believe, and show empirically, that the localization step is a missing link in the pipeline of object matching. As a starting point, the object of interest present in the moving image is located in the fixed image. This establishes an initial correspondence between the two images. It is different from the general purpose object detection scenario, where the system is expected to locate and identify each object present in the image. Thus, a modified version of single shot multi-box detector (SSD) \cite{liu2016ssd}, tweaked for the problem in hand, is used. Subsequently the masks corresponding to the detected objects are segmented and matched using spatial transformer network (STN) \cite{jaderberg2015spatial}. A smoothness constraint on the transformation parameters is included to prevent uncontrolled results. Contributions of this work are:
\begin{itemize}
	\item Introduction of object localization step in registration pipeline for making it suitable for real-world scenarios. In particular, the applications where considered images are not object-centric and populated by the objects of similar domains.
	
	\item Matching is performed using the segmented masks which eliminates intensity level differences and make the method suitable for multi-modal image registration.
	
	\item Being a learning based solution, the method is free from iterative optimization of a matching criteria and can be adapted for different tasks.
	
	\item Robustness of the method is demonstrated using three different dataset: MNIST, Caltech-101, and a ultrasound-CT dataset developed in this work.
	
\end{itemize}

\section{Related work}
Object matching and image registration are challenging problems. Traditional methods in this domain have focused on development of sophisticated descriptors and establish correspondence between key-points \cite{bay2006surf,harris1988combined,heinrich2012mind,lowe2004distinctive}. A major issue with these approaches is lack of generalization ability \cite{rocco2017convolutional}. Optimization of matching criteria with deformation field, displacement vector field, and diffeomorphic metric mapping is also a popular line work \cite{brox2011large,ashburner2007fast,ceritoglu2013computational,yang2016fast}. In addition, non-trainable optimization for object category matching have also been reported \cite{kim2013deformable,berg2005shape,duchenne2011graph}.

In contrast our approach is a combination of trainable CNNs which can be adapted to different modalities and tasks. We incorporate object localization step to handle large displacements and ambiguity arising from the surroundings. Some approaches have shown promising results with deep learning frameworks \cite{rocco2017convolutional,sokooti2017nonrigid,yang2017quicksilver}, however, they are trained with synthetic warps which often are insufficient for real-world domain of deformations. We do not rely on any such synthetic deformations. Further, unlike the recent approaches \cite{balakrishnan2018unsupervised,li2018non}, the proposed method does not rely on intensity level matching criteria and is suitable for multi-modal registration.

\section{Method}
The proposed method divides complex multi-modal object matching/registration problem into relatively simple sub-problems. Dedicated networks are employed to solve these sub-problems and combined to develop a novel pipeline. 

\subsection{Localization}
Localization step is the uniqueness of the proposed method. In contrast to many existing techniques, which register two or more images with assumption that the images contain only a single object, the proposed method is more generic and designed to handle images with multiple objects. A modified version of SSD is used for this purpose. Conventional SSD was designed to recognize each and every object in a single given image. On the other hand, the SSD in this work is simultaneously trained on moving as well as fixed images to identify the location of the moving image object in the fixed image. It establishes an initial correspondence between the two images without any knowledge of the object category or class label. Thus, instead of the original VGG-16 based architecture \cite{liu2016ssd}, a smaller architecture is used.
It is inspired from the work available at (\url{https://github.com/pierluigiferrari/ssd\_keras}) and shown in Table~\ref{tab:1}. The convolutional layers are followed by batch normalization and exponential
linear units and regularized using $L_2$ regularization with 0.0005 decay factor. Boxes of aspect ratios 0.5, 1.0 and 2.0 are generated from the outputs of the layers c4 to c7 at the scales varying from 0.08 to 0.96.

\begin{table}[!t]
	\centering
	\caption{SSD architecture.`c' and `p' represent the convolutional and pooling layers, respectively.}
	\label{tab:1}
	\begin{tabular}{|c|c|c|c|c|c|c|c|c|c|c|c|c|c|}
		\hline
		\bf Layer & c1 & p1 & c2 & p2 & c3 & p3 & c4 & p4 & c5 & p5 & c6 & p6 & c7 \\ \hline
		
		\bf Channels & 32 &-& 48&-&64&-&64&-&48&-&48&-&32\\ \hline
		\bf Kernel & 5$\times$5 & 2$\times$2& 3$\times$3& 2$\times$2& 3$\times$3& 2$\times$2& 3$\times$3& 2$\times$2& 3$\times$3& 2$\times$2& 3$\times$3& 2$\times$2& 3$\times$3\\ \hline
		\bf Stride & 1 &2& 1&2&1&2&1&2&1&2&1&2&1\\ \hline
		\bf Padding & 2 &0& 1&0&1&0&1&0&1&0&1&0&1\\ 
		\hline
	\end{tabular}
\end{table}
\begin{table}[!t]
	\centering
	\caption{STN architecture. `c', `p' and `FC' represent the convolutional, pooling and fully connected layers, respectively. The final layer is a TPS layer.}
	\label{tab:2}
	\begin{tabular}{|c|c|c|c|c|c|c|c|c|c|p{1cm}|p{1cm}|c|c|}
		\hline
		\bf Layer & c1 & c2 & p1 & c3 & c4 &p2 &c5&c6&p3 & FC$^*$& FC$^*$&TPS \\ \hline		
		\bf Channels & 32 &32 & -&64&64& - &128&128&-&256/ 1024/ 1024 &128/ 512/ 512 & 1\\ \hline
		\bf Kernel & 3$\times$3& 3$\times$3& 2$\times$2& 3$\times$3& 3$\times$3& 2$\times$2& 3$\times$3& 3$\times$3& 2$\times$2&-&-&-\\ \hline
		\bf Stride & 1 &1 & 2&1&1& 2 &1&1&2&- &- & -\\ \hline
		\bf Padding & 1 &1 & 0&1&1& 0 &1&1&0&- &- & -\\ \hline
		
	\end{tabular}
	
	$^*$ Channels used for MNIST/Caltech-101/US-CT experiments.
\end{table}
\subsection{Segmentation}
Localization step is followed by segmentation to map the intensity values of same object from different images to an identical levels. This brings the images on same manifold and makes it possible to perform multi-modal registration. Another advantage of segmentation step is the ground truth generation. In a real-world scenario, it is easier to annotate object segmentation on individual images as compared to marking the corresponding points in different images. Further, as noted in \cite{krause2015fine}, state-of-the-art segmentation networks can segment the images with accurate object boundaries. These approaches can be used to generate the masks corresponding to the object located in the previous step. However, to demonstrate the working of the proposed method manually marked ground truth segmentation are used during matching.



\subsection{Matching}
The final step of the proposed method is matching of the object masks. STN with thin plate spline (TPS) is used to perform non-rigid transformation of the object mask in the moving image. Although TPS is used in this work, the approach is not limited to it. Other transformations, for example, learnable bilinear interpolation or displacement vector field based transformation can also be used.

Similar to the localization step, a pair of object masks obtained from moving and fixed images are given to the STN, which learns the parameters of TPS for the desired matching. STN architecture used in this work is shown in Table~\ref{tab:2}. The convolutional layers are followed by leaky ReLU ($\alpha = 10^{-4}$) and regularized using $L_2$ regularization of weights with $10^{-4}$ decay factor. The final TPS layer\footnote{\url{https://github.com/mkturkcan/Spatial-Spline-Transformer-for-Keras/}} transforms the moving image mask to match it to the fixed image. The output of FC layer preceding TPS layer provides the parameters for TPS transformation. These parameters contain 6 global affine motion parameters and $2K$ coefficients (for 2D images) for displacement of the control points evenly spread on a 2D grid.

Dice coefficient (DC) based loss function is used to train the matching network. It is defined as
\begin{equation}
L_{DC} = 1 - 2 \times\frac{ \vert G_F \cap S_{M(\phi)}\vert}{ \vert G_F\vert +  \vert S_{M(\phi)}\vert}
\end{equation}
where $F$ and ${M(\phi)}$ are fixed and transformed moving images. $\phi$ represent the coefficients for the displacement of control points. $G_F$ and $S_{M(\phi)}$ represent the set of pixels belonging to the masks obtained from fixed and moving images, respectively. Further a smoothness constraint on the displacement of TPS control points is imposed to suppress any discontinuity occurring in the transformation. It is defined as 
\begin{equation}
L_{s} = \sum_{p=1}^{K}\left\|\nabla \phi (p)\right\|^2
\end{equation}
where $\nabla$ represent the spatial gradient corresponding to the grid of control points. The total loss is defined as
\begin{equation}
L = 1 - 2 \times\frac{ \vert G_F \cap S_{M(\phi)}\vert}{ \vert G_F\vert +  \vert S_{M(\phi)}\vert} + \lambda\sum_{p=1}^{K}\left\|\nabla \phi (p)\right\|^2
\end{equation}
where $\lambda$ can be considered as a regularization parameter. In this work a value of 1 is used for $\lambda$.

\section{Experiments and Results}

In order to evaluate the performance of our registration model, three different datasets are considered. Detection accuracy is measured as the overlap between detected and ground truth bounding boxes containing the object. Similarly matching accuracy is measured as overlap between the masks in fixed and moving images. DC and mean intersection over union (mIoU) are considered for this purpose.
\begin{equation}
\text{DC} = \frac{2\vert G \cap S\vert}{\vert G\vert + \vert S\vert}
\end{equation}
\begin{equation}
\text{mIoU} = \frac{\vert G \cap S\vert}{\vert G \cup S\vert}
\end{equation}
where $G$ and $S$ are the sets of the pixels representing detected and ground truth bounding boxes or masks of fixed and moving images, respectively.

As discussed, most existing methods ignore the detection part and work on the assumption that the images are object-centric which is not valid in general scenarios. In contrast deformable spatial pyramid (DSP) matching \cite{kim2013deformable}, one of the state-of-the-art methods, inherently localizes the object before performing registration. Therefore, we have evaluated the proposed method against DSP.

\subsection{Dataset Description}

First set of experiments is done on MNIST handwritten digits dataset. Each image in MNIST is of size 28$\times$28 pixels and contains only one digit. Images of same digits are considered for matching. MNIST digits differ from each other in terms writing style and stroke, therefore, form a reasonable dataset for matching. In order to understand the importance of localization, we generated images of size 112$\times$112 pixels, containing multiple digits and evaluated the performance of the proposed method. 

The second dataset is Caltech-101 dataset, which contains images of 101 categories. Images from different categories are divided into two sets, first, in which objects cover more than 50\% of the image region and second where object covers less then 50\% of image region. Moving and fixed images of an object are obtained from first and second set, respectively. Pairs of images generated from these sets are used for experiment.

The third dataset is a synthetic dataset which belongs to the medical image registration category. For this, ultrasound (US) images are simulated as in \cite{dillenseger2009fast} from CT images obtained from LiTS MICCAI 2017 challenge. Simulated data is used due to the unavailability of any public dataset for US-CT registration. A large sample from 2D CT scans containing liver and tumor regions are selected to generate US images. 

\subsection{MNIST Dataset experiments}
\begin{figure}[!t]
	\centering
	\begin{minipage}[]{0.4\linewidth}
		\centerline{\includegraphics[width=14cm]{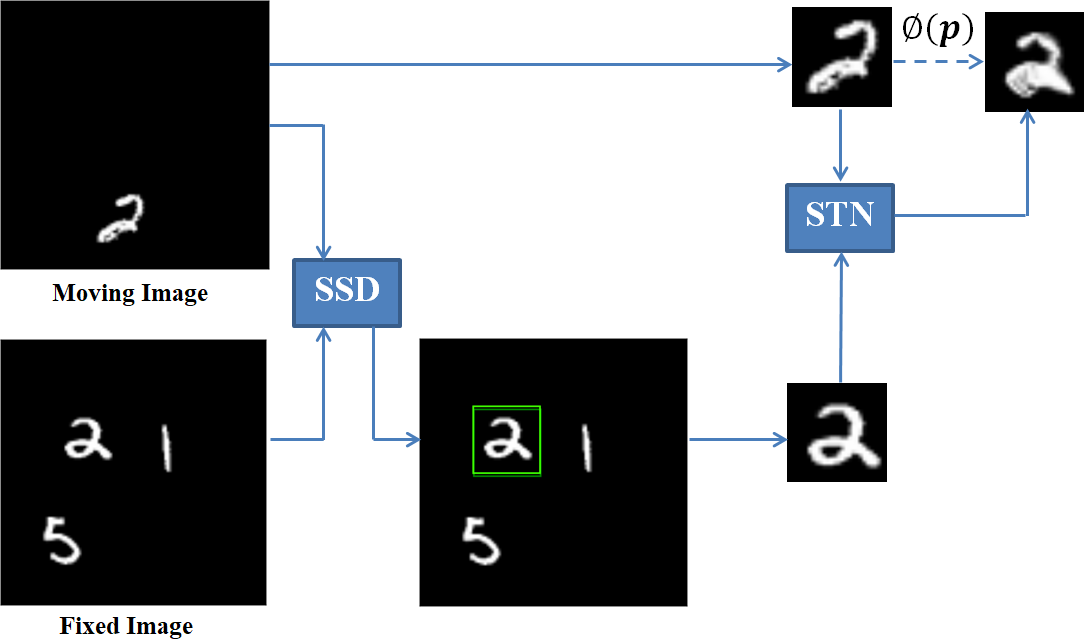}}
	\end{minipage}
	\caption[]{Image matching experiment on MNIST dataset.}
	
	\label{M}
\end{figure}

Training parameter values used in the proposed method for all experiments are listed in Table~\ref{tab:3}. First experiment is performed with MNIST dataset. Fig.~\ref{M} shows the execution flow of the proposed model on MNIST digits. Moving and fixed images are input to the model. In all samples, moving image contains a single digit, which is searched by the localization network in fixed image. Both the images are cropped and sent to STN for matching. MNIST contains binary images therefore does not require segmentation. Test set performance of the proposed method is summarized in Table~\ref{tab:4} and compared with DSP. 

\begin{table}[!t]
	\centering
	\caption{\emph{List of training parameters used in all experiments}} \label{tab:3}
	\begin{tabular}[l]{|>{\centering\arraybackslash}p{1.5cm}|>{\centering\arraybackslash}p{1cm}|>{\centering\arraybackslash}p{1cm}|>{\centering\arraybackslash}p{1cm}|>{\centering\arraybackslash}p{1cm}|>{\centering\arraybackslash}p{1cm}|>{\centering\arraybackslash}p{1cm}|}
		\hline
		\bf Parameter & \multicolumn{2}{c|}{\bf MNIST} & \multicolumn{2}{c|}{\bf Caltech101} & \multicolumn{2}{c|}{\bf US-CT}\\ \hline
		& {\bf SSD} & {\bf SNT}& {\bf SSD} & {\bf SNT}& {\bf SSD} & {\bf SNT}\\ \hline
		
		Dataset size (Train, Test, Val) & (100k, 10k, 10k)& (100k, 10k, 10k) & (15k, 5k, 5k) & (1725, 400, 400) & (1725, 400, 400)& (1725, 400, 400)\\ \hline
		Adam ($\beta_1$, $\beta_2$) & (0.9, 0.99) & (0.9, 0.99)& (0.9, 0.999) & (0.9, 0.99) & (0.9, 0.99)& (0.9, 0.99)\\ \hline
		Learning rate & 0.001 & 0.001& 0.001& 0.001& 0.001& 0.001\\ \hline
		Decay & $10^{-5}$ &$10^{-6}$&$10^{-5}$& $10^{-6}$& $10^{-5}$& $10^{-6}$\\ \hline
		Batch size & 32 & 128& 32& 8& 16& 8\\ \hline
		Epoch count & 400 & 500& 100& 100& 500& 100\\ \hline
		Input size & $112\times 112$&$28\times 28$&$320\times 320$&$224\times 224$&$512\times 512$&$256\times 256$\\
		\hline
		Control points ($p$) & -&64&-&256&-&256\\
		\hline
	\end{tabular}
\end{table}

\begin{figure}[!t]
	\centering
	\begin{minipage}[]{0.9\linewidth}
		\centerline{\includegraphics[width=11cm]{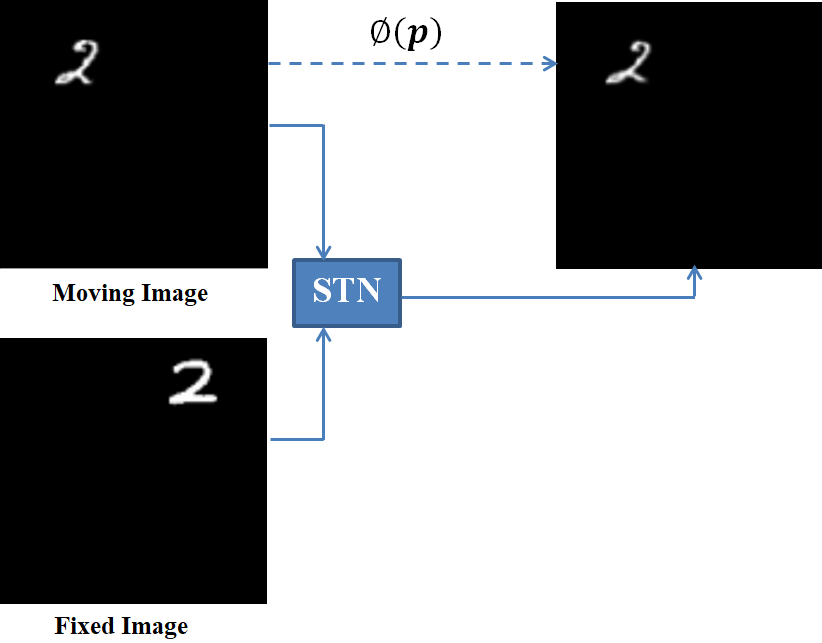}}
	\end{minipage}
	\caption[]{Image matching experiment without object localization on MNIST dataset.}
	
	\label{M_noSSD}
\end{figure}

\begin{table}[!t]
	\centering
	\caption{\emph{Observations on test images of MNIST dataset}} \label{tab:4}
	\begin{tabular}[l]{|c|c|c|}
		\hline
		\bf  Model/ Algorithm & {\bf DC} & {\bf mIoU }\\ \hline
		
		Detection Accuracy & 0.87 & 0.78\\ 
		(Proposed) & &\\ \hline
		Matching Accuracy & 0.79 & 0.66\\ 
		(Proposed) & &\\ \hline
		DSP acc w/o seg & 0.58 & 0.44\\ \hline
		
	\end{tabular}
\end{table}

MNIST dataset is further considered to understand the importance of localization step. Pairs of images of size 112$\times$112 are generated where each image contains a single digit. The digit images are randomly placed in 112$\times$112 pixel region. These pairs are used to train STN with 256 control points and rest of the parameters as listed in Table~\ref{tab:3}. Obtained result for a sample pair is included in Fig.~\ref{M_noSSD}. Clearly STN is unable to match the images in absence of localization.

\begin{figure*}[!t]
	\centering
	\begin{minipage}[]{0.4\linewidth}
		\centerline{\includegraphics[width=15cm]{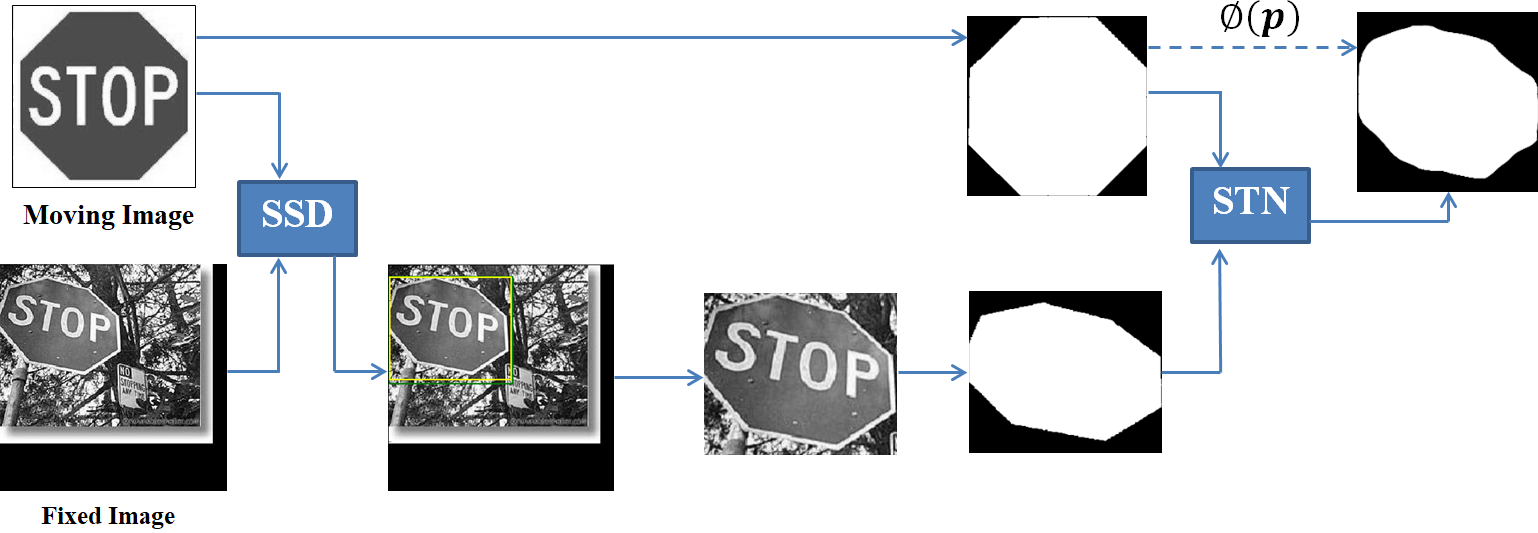}}
	\end{minipage}
	\caption[]{Image matching experiment on Caltech-101 dataset.}
	
	\label{C}
\end{figure*}

\subsection{Caltech Dataset experiments}
In this experiment, samples from the set in which objects cover less than 50\% of image regions are considered as fixed images and the samples from the other set are considered as moving images. To demonstrate the working of the proposed method, we have considered ground truth masks of the objects, however for real applications, equivalent masks can be generated using state-of-the-art segmentation techniques.  

Localization network is able to detect and create bounding box around the desired object in fixed image, as shown in Fig.~\ref{C}. Image matching part is done on the masks of moving and fixed images. The test set results are listed in Table~\ref{tab:5}. Similar to MNIST, the proposed model is able to outperform DSP in Caltech-101 dataset too.

\begin{figure}[!t]
	\centering
	\begin{minipage}[]{0.4\linewidth}
		\centerline{\includegraphics[width=8cm]{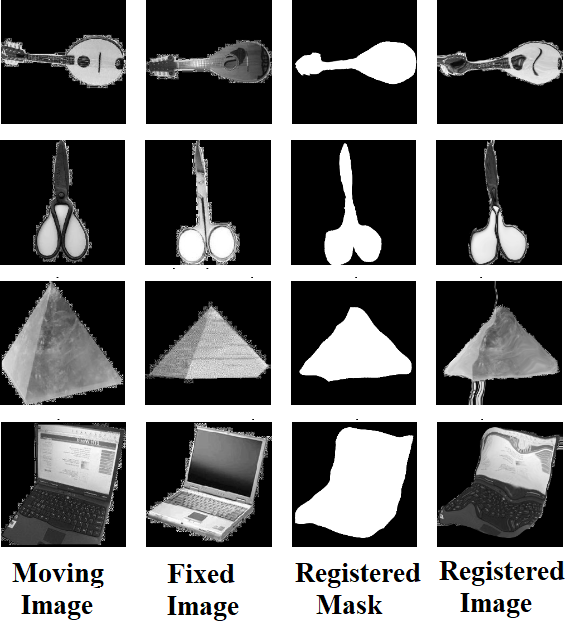}}
	\end{minipage}
	\caption[]{Image matching experiment on Caltech-101 dataset. Masks without semantically distinguished parts are used for training and testing.}
	\label{C1}
\end{figure}
\begin{table}[!t]
	\centering
	\caption{\emph{Observations on test images Caltech-101 dataset}} \label{tab:5}
	\begin{tabular}[l]{|c|c|c|}
		\hline
		\bf  Model/ Algorithm & {\bf DC} & {\bf mIoU }\\ \hline
		
		Detection Accuracy & 0.95 & 0.91 \\  
		(Proposed) & &\\ \hline
		Matching Accuracy & 0.94 & 0.90\\  
		(Proposed) & &\\ \hline
		DSP acc w/o seg & 0.60 & 0.48 \\ \hline
		
	\end{tabular}
\end{table}

\begin{figure}[!t]
	\centering
	\begin{minipage}[]{0.4\linewidth}
		\centerline{\includegraphics[width=8cm]{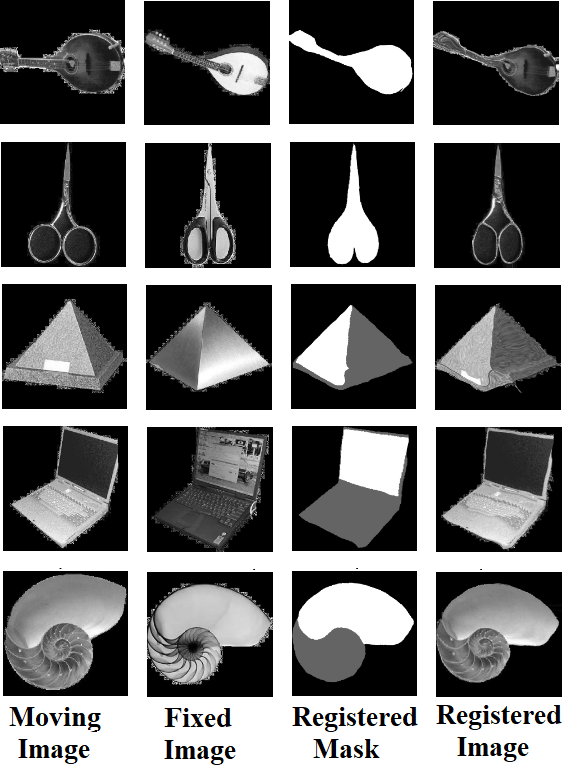}}
	\end{minipage}
	\caption[]{Image matching experiment on Caltech-101 dataset. Masks with at most two semantically distinguished parts are used for training and testing.}
	\label{C2}
\end{figure}

\begin{figure*}[!t]
	\centering
	\begin{minipage}[]{0.4\linewidth}
		\centerline{\includegraphics[width=15cm]{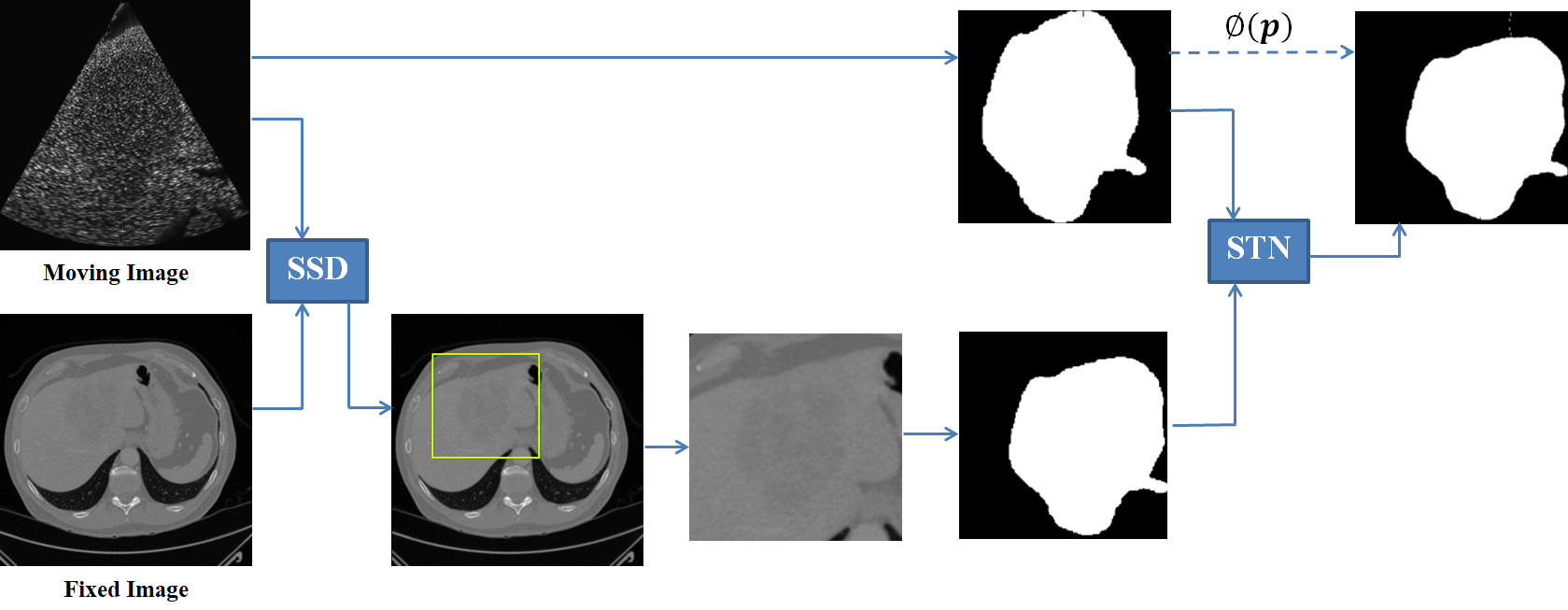}}
	\end{minipage}
	\caption[]{Image matching experiment on US-CT dataset.}
	
	\label{U}
	
\end{figure*}
For further analysis, the TPS transformation with, parameter predicted by STN during the registration of the segmented masks is applied on moving image and the outputs are shown in Fig.~\ref{C1}. As can be seen, the model is able to preserve object boundaries during transformation but not able to preserve internal details. Note that the matching is performed at the shape level using the masks which do not contain any information about semantically distinguishable parts of the object. For better understanding, another experiment is conducted where STN is trained using masks which included different labels for different parts of the object. The number of labels are restricted to two and mask is at most divided into two parts. Observations on test images are shown in Fig.~\ref{C2}. A considerable improvement in the quality of the outputs is observed. The improvement can be attributed to the learned notion of internal edges. This suggests that an accurate image matching can be obtained by incorporating fine part annotations in the masks or by training STN with edge maps.

\subsection{US and CT Image Dataset}
Final experiments are performed with US and CT dataset. The simulated US images are considered as moving images while CT images are used as fixed images. Our localization network is efficiently able to identify regions in CT images corresponding to the US images. This helps in improving the overall accuracy of the registration model. Tumors are considered as the objects for registration. As ground truth segmentation of LiTS CT images are available, the intermediate masks are directly obtained from the ground truths. Matching performance of the proposed model is evident from Fig.~\ref{U}. Test set observations are summarized in Table~\ref{tab:6}. DSP is originally applied on US and CT images itself which results in comparatively lower performance. Later it is tested with tumor masks and a considerable gain in the performance is observed, which explains the better performance of the proposed approach.

\begin{table}[!t]
	\centering
	\caption{\emph{Observations on test images of US-CT dataset}} \label{tab:6}
	\begin{tabular}[l]{|c|c|c|}
		\hline
		\bf  Model/ Algorithm & {\bf DC} & {\bf mIoU }\\ \hline
		
		Detection Accuracy & 0.97 & .0.94\\  
		(Proposed) & &\\ \hline
		Matching Accuracy &  0.96 & 0.94\\  
		(Proposed) & &\\ \hline
		DSP acc w/o seg & 0.29  & 0.17\\ \hline
		DSP acc with seg & 0.63 & 0.49\\ \hline
		
	\end{tabular}
\end{table}
\section{Conclusion}
A novel pipeline of CNNs is proposed in this work to perform multi-modal object matching and registration. The complex registration problem is divided into comparatively simple sub-problems of object localization, mask generation using segmentation and mask matching. It gives promising results in MNIST, Caltech-101 and US-CT dataset which shows that the proposed method is suitable for the desired objective.

\bibliography{paper}

\begin{thebibliography}{10}
\providecommand{\url}[1]{#1}
\csname url@samestyle\endcsname
\providecommand{\newblock}{\relax}
\providecommand{\bibinfo}[2]{#2}
\providecommand{\BIBentrySTDinterwordspacing}{\spaceskip=0pt\relax}
\providecommand{\BIBentryALTinterwordstretchfactor}{4}
\providecommand{\BIBentryALTinterwordspacing}{\spaceskip=\fontdimen2\font plus
\BIBentryALTinterwordstretchfactor\fontdimen3\font minus
  \fontdimen4\font\relax}
\providecommand{\BIBforeignlanguage}[2]{{%
\expandafter\ifx\csname l@#1\endcsname\relax
\typeout{** WARNING: IEEEtran.bst: No hyphenation pattern has been}%
\typeout{** loaded for the language `#1'. Using the pattern for}%
\typeout{** the default language instead.}%
\else
\language=\csname l@#1\endcsname
\fi
#2}}
\providecommand{\BIBdecl}{\relax}
\BIBdecl

\bibitem{barnes2010generalized}
C.~Barnes, E.~Shechtman, D.~B. Goldman, and A.~Finkelstein, ``The generalized
  patchmatch correspondence algorithm,'' \emph{European Conference on Computer
  Vision}, pp. 29--43, 2010.

\bibitem{nam2011automatic}
W.~H. Nam, D.-G. Kang, D.~Lee, J.~Y. Lee, and J.~B. Ra, ``Automatic
  registration between 3d intra-operative ultrasound and pre-operative ct
  images of the liver based on robust edge matching,'' \emph{Phys. Med. Biol.},
  vol.~57, no.~1, pp. 69--91, 2011.

\bibitem{ceritoglu2013computational}
C.~Ceritoglu, X.~Tang, M.~Chow, D.~Hadjiabadi, D.~Shah, T.~Brown, M.~H.
  Burhanullah, H.~Trinh, J.~Hsu, K.~A. Ament \emph{et~al.}, ``Computational
  analysis of lddmm for brain mapping,'' \emph{Frontiers in neuroscience},
  vol.~7, p. 151, 2013.

\bibitem{ashburner2007fast}
J.~Ashburner, ``A fast diffeomorphic image registration algorithm,''
  \emph{Neuroimage}, vol.~38, no.~1, pp. 95--113, 2007.

\bibitem{revaud2016deepmatching}
J.~Revaud, P.~Weinzaepfel, Z.~Harchaoui, and C.~Schmid, ``Deepmatching:
  Hierarchical deformable dense matching,'' \emph{International Journal of
  Computer Vision}, vol. 120, no.~3, pp. 300--323, 2016.

\bibitem{rohe2017svf}
M.-M. Roh{\'e}, M.~Datar, T.~Heimann, M.~Sermesant, and X.~Pennec, ``Svf-net:
  Learning deformable image registration using shape matching,''
  \emph{International Conference on Medical Image Computing and
  Computer-Assisted Intervention}, pp. 266--274, 2017.

\bibitem{rocco2017convolutional}
I.~Rocco, R.~Arandjelovic, and J.~Sivic, ``Convolutional neural network
  architecture for geometric matching,'' \emph{Proc. CVPR}, vol.~2, 2017.

\bibitem{liu2016ssd}
W.~Liu, D.~Anguelov, D.~Erhan, C.~Szegedy, S.~Reed, C.-Y. Fu, and A.~C. Berg,
  ``{SSD}: Single shot multibox detector,'' \emph{European conference on
  computer vision}, pp. 21--37, 2016.

\bibitem{jaderberg2015spatial}
M.~Jaderberg, K.~Simonyan, A.~Zisserman \emph{et~al.}, ``Spatial transformer
  networks,'' \emph{Advances in neural information processing systems}, pp.
  2017--2025, 2015.

\bibitem{bay2006surf}
H.~Bay, T.~Tuytelaars, and L.~Van~Gool, ``Surf: Speeded up robust features,''
  \emph{European conference on computer vision}, pp. 404--417, 2006.

\bibitem{harris1988combined}
C.~Harris and M.~Stephens, ``A combined corner and edge detector.'' \emph{Alvey
  vision conference}, vol.~15, no.~50, pp. 10--5244, 1988.

\bibitem{heinrich2012mind}
M.~P. Heinrich, M.~Jenkinson, M.~Bhushan, T.~Matin, F.~V. Gleeson, M.~Brady,
  and J.~A. Schnabel, ``{MIND}: Modality independent neighbourhood descriptor
  for multi-modal deformable registration,'' \emph{Medical image analysis},
  vol.~16, no.~7, pp. 1423--1435, 2012.

\bibitem{lowe2004distinctive}
D.~G. Lowe, ``Distinctive image features from scale-invariant keypoints,''
  \emph{International journal of computer vision}, vol.~60, no.~2, pp. 91--110,
  2004.

\bibitem{brox2011large}
T.~Brox and J.~Malik, ``Large displacement optical flow: descriptor matching in
  variational motion estimation,'' \emph{IEEE transactions on pattern analysis
  and machine intelligence}, vol.~33, no.~3, pp. 500--513, 2011.

\bibitem{yang2016fast}
X.~Yang, R.~Kwitt, and M.~Niethammer, ``Fast predictive image registration,''
  \emph{Deep Learning and Data Labeling for Medical Applications}, pp. 48--57,
  2016.

\bibitem{kim2013deformable}
J.~Kim, C.~Liu, F.~Sha, and K.~Grauman, ``Deformable spatial pyramid matching
  for fast dense correspondences,'' \emph{Proceedings of the IEEE Conference on
  Computer Vision and Pattern Recognition}, pp. 2307--2314, 2013.

\bibitem{berg2005shape}
A.~C. Berg, T.~L. Berg, and J.~Malik, ``Shape matching and object recognition
  using low distortion correspondences,'' \emph{Computer Vision and Pattern
  Recognition, 2005. CVPR 2005. IEEE Computer Society Conference on}, vol.~1,
  pp. 26--33, 2005.

\bibitem{duchenne2011graph}
O.~Duchenne, A.~Joulin, and J.~Ponce, ``A graph-matching kernel for object
  categorization,'' \emph{Computer Vision (ICCV), 2011 IEEE International
  Conference on}, pp. 1792--1799, 2011.

\bibitem{sokooti2017nonrigid}
H.~Sokooti, B.~de~Vos, F.~Berendsen, B.~P. Lelieveldt, I.~I{\v{s}}gum, and
  M.~Staring, ``Nonrigid image registration using multi-scale 3d convolutional
  neural networks,'' \emph{International Conference on Medical Image Computing
  and Computer-Assisted Intervention}, pp. 232--239, 2017.

\bibitem{yang2017quicksilver}
X.~Yang, R.~Kwitt, M.~Styner, and M.~Niethammer, ``Quicksilver: Fast predictive
  image registration--a deep learning approach,'' \emph{NeuroImage}, vol. 158,
  pp. 378--396, 2017.

\bibitem{balakrishnan2018unsupervised}
G.~Balakrishnan, A.~Zhao, M.~R. Sabuncu, J.~Guttag, and A.~V. Dalca, ``An
  unsupervised learning model for deformable medical image registration,''
  \emph{Proceedings of the IEEE Conference on Computer Vision and Pattern
  Recognition}, pp. 9252--9260, 2018.

\bibitem{li2018non}
H.~Li and Y.~Fan, ``Non-rigid image registration using self-supervised fully
  convolutional networks without training data,'' \emph{Biomedical Imaging
  (ISBI 2018), 2018 IEEE 15th International Symposium on}, pp. 1075--1078,
  2018.

\bibitem{krause2015fine}
J.~Krause, H.~Jin, J.~Yang, and L.~Fei-Fei, ``Fine-grained recognition without
  part annotations,'' \emph{Proceedings of the IEEE Conference on Computer
  Vision and Pattern Recognition}, pp. 5546--5555, 2015.

\bibitem{dillenseger2009fast}
J.-L. Dillenseger, S.~Laguitton, and {\'E}.~Delabrousse, ``Fast simulation of
  ultrasound images from a ct volume,'' \emph{Computers in biology and
  medicine}, vol.~39, no.~2, pp. 180--186, 2009.

\end{thebibliography}
\bibliographystyle{IEEEtran}

\end{document}